\pdfoutput=1

\documentclass[11pt]{article}

\usepackage[]{EMNLP2023}

\usepackage{times}
\usepackage{latexsym}

\usepackage[T1]{fontenc}

\usepackage[utf8]{inputenc}

\usepackage{microtype}

\usepackage{inconsolata}
\usepackage{textcomp}
\usepackage{stfloats}
\usepackage{url}
\usepackage{color}
\usepackage{verbatim}
\usepackage{graphicx}

\usepackage{multicol}
\usepackage{multirow}
\usepackage{hhline}
\usepackage{makecell}
\usepackage{bm}
\usepackage{xcolor}

\usepackage{bbm} 
\usepackage{enumerate}
\usepackage{enumitem}
\usepackage{amsmath}

\usepackage{booktabs}

\usepackage{stfloats}
\usepackage{CJKutf8}
\usepackage[normalem]{ulem}
\useunder{\uline}{\ul}{}

\usepackage{footnote}
\usepackage{caption}
\usepackage{etoolbox}

\usepackage{amssymb}
\usepackage{algorithm}
\usepackage{algorithmic}

\newcommand{\ie}{\emph{i.e., }}
\newcommand{\eg}{\emph{e.g., }}

\title{Knowledge-enhanced Memory Model for Emotional Support Conversation}

\author{Mengzhao Jia$^1$, {\bf Qianglong Chen}$^2$, {\bf Liqiang Jing}$^3$, {\bf Dawei Fu}$^4$, {\bf Renyu Li}$^4$\thanks{~~Renyu Li is the corresponding author.} \\   $^1$University of Notre Dame\\ $^2$Zhejiang University  \\$^3$University of Texas at Dallas \\$^4$Alibaba Group  \\ jiamengzhao98@gmail.com, chenqianglong@zju.edu.cn, jingliqiang6@gmail.com,\\ \{dawei.fdw, renyu.rl\}@alibaba-inc.com}

\begin{document}
\maketitle
\begin{abstract}
The prevalence of mental disorders has become a significant issue, leading to the increased focus on Emotional Support Conversation as an effective supplement for mental health support. Existing methods have achieved compelling results, however, they still face three challenges: 1) variability of emotions, 2) practicality of the response, and 3) intricate strategy modeling. To address these challenges, we propose a novel knowledge-enhanced Memory mODEl for emotional suppoRt coNversation (\textbf{MODERN}). Specifically, we first devise a knowledge-enriched dialogue context encoding to perceive the dynamic emotion change of different periods of the conversation for coherent user state modeling and select context-related concepts from ConceptNet for practical response generation. Thereafter, we implement a novel memory-enhanced strategy modeling module to model the semantic patterns behind the strategy categories. Extensive experiments on a widely used large-scale dataset verify the superiority of our model over cutting-edge baselines. 
\end{abstract}


\section{Introduction}

\begin{figure}
    \centering
    \includegraphics[width=0.85\linewidth]{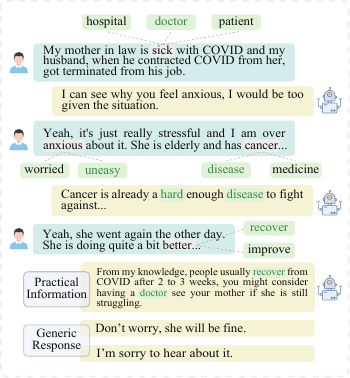}
    \caption{Illustration of an emotional support conversation example. The words with a green background are the retrieved concepts from ConceptNet~\cite{DBLP:conf/aaai/SpeerCH17}. 
    }
    \label{fig:intro}
\end{figure}
Mental disorders are known for their high burden, with more than 50\% of adults experiencing a mental illness or disorder at some point in their lives; yet despite its high prevalence, only one in five patients receive professional treatment\footnote{\url{https://tinyurl.com/4r8svsst}.}. 
Recent studies have shown that an effective mental health intervention method is the provision of emotional support conversations~\cite{green2022emotional,chang2021relationships}. As such, Emotional Support Conversations (ESConv), as defined by~\citet{DBLP:conf/acl/LiuZDSLYJH20}, has garnered substantial attention in recent years. They have emerged as a promising alternative strategy for mental health intervention, paving the way for the development of neural dialogue systems designed to provide support for those in need. 

The ESConv takes place between a help-seeker (user) and a supporter (dialogue model) in a multi-turn manner. 
It requires the dialogue model to employ a range of supportive strategies effectively, easing the emotional distress of the users and helping them overcome the challenges they face. 
Prior research primarily concentrated on two aspects. The first aimed to enhance the model's comprehension of the contextual semantics in the conversations, such as the user's situation, emotions, and intentions. An example of these efforts is the work of \citet{DBLP:conf/ijcai/00080XXSL22} who designed a hierarchical graph network to capture the overall emotional problem cause and specific user intentions.
The second aspect focused on predicting the dialogue strategy accurately and responding based on the predicted strategy category. For example, \citet{DBLP:conf/emnlp/ChengLLWZLL022} employed a lookahead heuristics for dialogue strategy planning and selection.


%


Despite the success of existing studies, this task is non-trivial due to the following three challenges. 
1) \textbf{Variability of emotions.} As the conversation progresses, the user's emotional state evolves subtly and constantly. Accurately recognizing the emotional change is indispensable to understanding the user's real-time state and thus responding empathically \cite{Ding2022, Burleson2003, HighSteuber2014}. How to model the dynamic emotional change during the dialogue process is the first challenge. 
2) \textbf{Practicality of the response.} In the absence of explicit cues, neural dialogue systems are inclined to make generic responses~\cite{DBLP:conf/icassp/WeiLMZPLJ19,liu-etal-2018-towards-less}. As shown in Figure~\ref{fig:intro}, the generic responses are deficient to provide personalized and suitable suggestions for the user's specific concerns. To resolve this issue, introducing context-related concepts (\eg \textit{doctor} and \textit{recover}) can promote generating more meaningful and actionable suggestions for specific situations. As a result, the integration of appropriate concepts poses a non-trivial challenge in generating practical responses. 
3) \textbf{Intricate strategy modeling.} Dialogue strategy, as a kind of linguistic pattern, has been reported as a highly complex concept encompassing various intricate linguistic features~\cite{hill2009helping, zheng-etal-2021-comae}. Previous work resorted to a single vector (\ie a category indicator) for strategy representation, which is insufficient to fully represent the complex strategy pattern information. Therefore, how to model strategy information sufficiently is the third challenge.

To overcome these challenges, as shown in Figure~\ref{fig:model}, we introduce a novel knowledge-enhanced Memory mODEl for emotional suppoRt coNversation, dubbed MODERN. 
In particular, MODERN adapts the BART~\cite{DBLP:conf/acl/LewisLGGMLSZ20} as its backbone and consists of the knowledge-enriched dialogue context encoding module, the memory-enhanced strategy modeling module, and the response decoding module. 
To capture the emotional change as the conversation progresses, the first module detects the emotions for all utterances and explicitly injects them into the dialogue context as a kind of emotional knowledge, thus understanding the user’s status coherently. 
In addition, this module also introduces the concepts reasoning and selection component to acquire valid context-related concepts from a knowledge graph called ConceptNet~\cite{DBLP:conf/aaai/SpeerCH17} and incorporate them into the dialogue context to fulfill meaningful and practical suggestion generation. 
Moreover, in contrast to existing studies that depend on simplistic indicators to represent strategy categories, the memory-enhanced strategy modeling module learns strategy patterns by a strategy-specific memory bank. In this way, it can detect and mimic the intricate patterns in human emotional support strategies.  
Finally, the third module aims to generate the target response with the BART decoder. 

Our contributions can be summarized as follows: 
1) We first analyze the current challenges of the ESConv task, and according to that propose a novel knowledge-enhanced Memory mODEl for emotional suppoRt coNversation, named MODERN, which can model complex supportive strategy as well as utilize emotional knowledge and context-related concepts to perceive the variability of emotions and provide practical support advice. 
2) We propose a memory-enhanced strategy modeling module, where a unique memory bank is designed to model intricate strategy patterns, and an auxiliary strategy classification task is introduced to distill the strategy pattern information. 
3) We present a thorough validation and evaluation of our model, providing an in-depth analysis of the results and a comparison with other models. 
The extensive experiments on the ESConv dataset~\cite{DBLP:conf/acl/LiuZDSLYJH20} demonstrate that MODERN achieves state-of-the-art performance under both automatic and human evaluations. 

\section{Related Work}
\textbf{Emotional and Empathetic Dialogue Systems}.  
With the popularity and growing success of dialogue systems, many research interests have recently endeavored to empower the system to reply with a specific and proper emotion, therefore forming a more human-like conversation. Particularly, two research directions arise researchers’ interest, namely the emotional~\cite{DBLP:conf/aaai/ZhouHZZL18} and empathetic~\cite{DBLP:conf/acl/RashkinSLB19} response generation. The former direction expects the dialogue agent to respond with a given emotion~\cite{DBLP:conf/acl/ShenF20,DBLP:conf/naacl/ChenLY22,DBLP:conf/naacl/KimAKL22,DBLP:conf/coling/LiCRRTC20}.
While the latter requires the dialogue system to actively detect and understand the users' emotions and then respond with an appropriate emotion~\cite{DBLP:conf/emnlp/MajumderHPLGGMP20}. For example, \citeauthor{DBLP:conf/emnlp/LinMSXF19} utilized multiple decoders as different listeners to react to different emotions and then softly combine the output states of the decoders appropriately based on the recognized user's emotion. 
Nevertheless, unlike above directions, the ESConv task concentrates on alleviating users' negative emotion intensity and providing supportive instructions to help them overcome struggling situations. 

\noindent \textbf{Emotional Support Conversation}. 
As an emerging research task, emotional support conversation has gradually attracted intense attention in recent years. Existing works mostly focus on two aspects. The first is to understand the complicated user emotions and intentions in the dialogue context. Specifically, they explored context semantic relationships~\cite{DBLP:journals/corr/abs-2210-12640,DBLP:conf/ijcai/00080XXSL22}, commonsense knowledge~\cite{DBLP:conf/acl/TuLC0W022,DBLP:conf/ijcai/00080XXSL22}, or emotion causes~\cite{DBLP:conf/emnlp/ChengLLWZLL022} to better capture and understand the emotions and intentions of users. The other trend in addressing the task is to predict the strategy category accurately so as to respond in accordance with it~\cite{DBLP:journals/kbs/PengQHXL23,DBLP:journals/corr/abs-2210-12640}. For instance,~\citeauthor{DBLP:conf/acl/TuLC0W022} firstly proposed to predict a strategy probability distribution and generate the response guided by a mixture of multiple strategies. Despite their remarkable achievements, existing work still face three challenges: intricate strategy modeling, variability of emotions, and practicality of the responses.


\section{Task Formulation}

\begin{figure*}
    \centering
    \includegraphics[width=\textwidth]{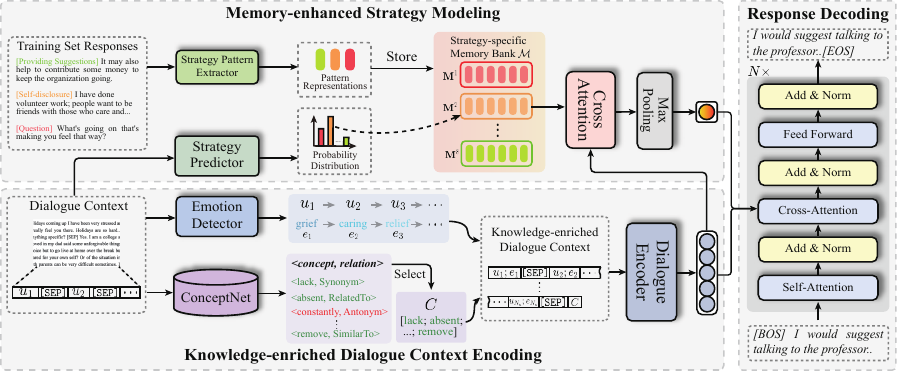}
    \caption{Illustration of the proposed MODERN framework, which consists of three key components: Strategy Memory-enhanced Dialogue Context Encoding, Multi-source Knowledge Injection, and Response Decoding.}
    \label{fig:model}
\end{figure*}



In the setting of emotional support conversation, suppose we have a training dataset $\mathcal{P}=\{p_1,p_2,\dots,p_V\}$ composed of $V$ samples. Each sample $p_i = \{t_i,\mathcal{D}_i,g_i,R_i\}$ includes a seeker's situation $t_i$, a dialogue context $\mathcal{D}_i$, a support strategy $g_i$, and a target response $R_i$. Therein, $\mathcal{D}_i$ contains a sequence of ${N}_{u}^{i}$ history utterances between the user and the supporter, denoted as $\mathcal{D}_i=(u_1^{i}, u_2^{i}, \dots, u^i_{{N}_{u}^{i}})$. $R_i=(r^i_1, r^i_2, \dots, r^i_{N_{r}^{i}})$ is the supportive response with ${N_{r}^{i}}$ tokens. 
The goal of the ESConv task is to learn a model $\mathcal{F}$ that can generate a supportive response $\hat{R}_i$ referring to the input context $\mathcal{D}_i$ and situation $t_i$ as follows,
\begin{equation}
    \hat{R}_i = \mathcal{F}(\mathcal{D}_i, t_i|\Theta),
\end{equation}
where $\Theta$ refers to the set of to-be-learned parameters of the model $\mathcal{F}$. Notably, $g_i$ can also be utilized in the model training stage but is not available in the inference phrase. 
For brevity, we temporally omit the superscript $i$ of the $i$-th sample in the rest of this paper.
  




\section{Method}
In this section, we detail the proposed model MODERN demonstrated in Figure~\ref{fig:model}. 

\subsection{Knowledge-enriched Dialogue Context Encoding}
In this section, we first utilize an emotion detector to recognize fine-grained emotions of utterances as emotional knowledge for capturing emotional change. In addition, we select related concepts from the ConceptNet knowledge graph for meaningful and practical suggestion generation. 

\textbf{Change-aware Emotion Detection}
Psychiatric and mental health studies have proved that empathy is essential to emotionally helping relationships~\cite{reynolds1999empathy,liu2018should,parkin2014greater}. And fine-grained emotional information is one of the key factors to enhance the empathetic ability~\cite{DBLP:conf/aaai/LiLRRC22}. Apart from the static emotional signals, dynamic emotional changes during the conversation progress are also beneficial. Concretely, change-aware emotion information enriches the model to understand the user's status coherently. 
Inspired by this, we devise to identify the user's fine-grained emotions and perceive the dynamic changes of emotions in the dialogue context.

Specifically, we start by obtaining the fine-grained emotion via an off-the-shelf pretrained emotion detector\footnote{\url{https://huggingface.co/arpanghoshal/EmoRoBERTa}.}, which can recognize up to 28 different emotional categories. We apply the detector to every utterance in the dialogue context for emotion recognition as follows,
\begin{equation}
    e_j = \operatorname{Emo}(u_j), \quad 0\leq j \leq N_u,
\end{equation}
where $e_j$ is the predicted emotional category word representing the detected emotion in $u_j$. Thereafter, we directly inject the natural language form of emotion category words into the dialogue context as additional emotion knowledge. This practice aligns closely with the input format of the pretrained BART model. 
Moreover, it also avoids introducing unnecessary parameters that would interfere with the generative model learning. Empirically, we concatenate the emotional category into the sequence of dialogue context tokens, denoted as $ I = [u_1;e_1; \mathtt{SEP}; \dots u_j;e_j; \mathtt{SEP}; \dots  u_{N_u};e_{N_u}]$. In this way, the dynamic emotional changes corresponding to the dialogue progress can be coherently exploit by the emotinal support model. 

\textbf{Context-related Concepts Reasoning and Selection}. 
Considering ConceptNet involve abundant general human knowledge, which plays an important role in understanding human situations and associating them with practical suggestions, we select potentially useful context-related concepts to enrich the model to generate responses with high informativeness. 
Concretely, the ConceptNet knowledge graph involves 3.1 million concepts and 38 million relations, which can be used to mine the underlying concept information in the dialogue context. 
We first identify all the concepts in ConceptNet that are mentioned in the dialogue context and remove the top-$K$ frequent concepts in the training set because these words are usually too general to provide valid suggestions for a specific situation, such as \textit{help}, \textit{thing}, and \textit{feeling}. In this way, we derive $N_c$ concepts, denoted as $\{c_1, \cdots, c_{N_c}\}$.

Thereafter, we leverage the $N_c$ concepts as the anchors to reason the related concepts. Specifically, for each anchor concept $c$, we retrieve all its one-hop neighboring concept-relation pairs from the ConceptNet.  
Mathematically, let $\mathcal{N}(c)$ be the set of neighboring concept-relation pairs of the anchor concept $c$. Then the context-related concept-relation pairs can be represented as $\{\mathcal{N}(c_1),\mathcal{N}(c_2), \cdots, \mathcal{N}(c_{N_c})\}$, where $\mathcal{N}({c_i})=\{(a^1_{c_1},h^1_{c_1}),\cdots,(a^{N_{c_i}}_{c_1},h^{N_{c_i}}_{c_1})\}$ is a set of $N_{c_i}$ retrieved $<concept,relation>$ pairs. 
Following~\citet{DBLP:conf/iwsds/Liu0MU21}, to avoid introducing extraneous concepts, we filter out concepts with \textit{excluded} relations, \eg \textit{Antonym}, \textit{ExternalURL}, and \textit{NotCapableOf}. 
Finally, we concatenate the concepts in the filtered pairs and deem them as context-related concepts $C$. 

\textbf{Knowledge-enriched Dialogue Context Embedding}. To acquire the representation of the dialogue context and corresponding knowledge, we first encode the dialogue context sequence (user situation $t$, emotional-aware dialogue context $I$, and concepts $C$) with a Transformer-based encoder as follows,
\begin{equation}
    \mathbf{H} = \operatorname{Enc}_c([t;I;C]),
\end{equation}
where $\mathbf{H} \in \mathbb{R}^{L \times d}$ 
denotes the hidden contextual representation. $L$ and $d$ refer to the number of tokens in $[t;I;C]$ and the representation hidden dimension, respectively.



\subsection{Memory-enhanced Strategy Modeling}
During the conversation, the supporter adopts different strategies for different purposes, ultimately achieving the goal of reducing the intensity of the user's negative emotions. For example, using the \textit{Question} strategy helps the supporter to explore the concrete situation faced by the user, while the use of the \textit{Reflection of Feelings} strategy conveys the supporter's understanding of the user's current emotions. Existing work constrains the model to responding under a strategy category by simply providing a single vector indicating the strategy's name or description. However, the semantic patterns of strategies are highly complex, the name or description is not able to contain the specific linguistic patterns (\eg expression manner, wording, and phrasing) of the strategy. Therefore, inspired by~\cite{stylized}, we propose to disentangle the strategy patterns from the same-strategy responses to provide more specific guidance for the strategy-constrained generation. 

\textbf{Strategy Pattern Modeling}. 
We first acquire strategy pattern representations of each responses in the training set via a strategy pattern extractor $\operatorname{Enc}_r$ as follows,
\begin{equation}
    \mathbf{r} = \operatorname{MaxPooling}(\operatorname{Enc}_r({R})), 
\end{equation}
where $\mathbf{r} \in \mathbb{R}^d$ denotes the strategy pattern representation. 
Meanwhile, in order to accurately capture the strategy pattern information and avoid irrelevant disturbance, we design an auxiliary strategy classification task to guide the extractor to map more strategy-related information into the pattern representation. 
The auxiliary objective $\mathcal{L}_r$ is derived by the following loss function, 
\begin{equation}
    \mathcal{L}_r = -\log {p}\left({g}|\mathbf{r} \right).
\end{equation}

\textbf{Strategy-specific Memory Bank}. To utilize as much strategy pattern information as possible, instead of using a single representation vector, we devise a memory bank mechanism to store multiple pattern representations according to their strategy categories. 
In particular, we denote the memory bank as $\mathcal{M}= \{\mathbf{M}^{1},\dots,\mathbf{M}^{G}\}$, in which $\mathbf{M}^{g} \in \mathbb{R}^{N_s^g \times d}$ is a matrix of $N_s^g$ pattern representations corresponding to $g$-th strategy category, and $G$ is the total number of strategy categories. $N_s^g$ is $0$ at the initial training step. As the training progresses, $N_s^g$ continues to increase until the maximum threshold $N_m$ is reached. In particular, we store pattern representation of $g$-th strategy category into the corresponding $\mathbf{M}^{g}$ by concatenation as follows, 
\begin{equation}
    \mathbf{{M}}^g \longleftarrow [\mathbf{{M}}^g; \mathbf{r}_{g}], \label{add}
\end{equation}
where $\mathbf{r}_{g}$ denotes a representation belongs to the $g$-th strategy category and $[\cdot;\cdot]$ refers to the concatenation operation. As the representations are optimized along with the classification training process, 
we dynamically update $\mathbf{M}^{g}$ in a \textit{first-in-first-out} manner as follows,
\begin{equation}
\label{update}
    \mathbf{M}^{g} =  \left\{
    \begin{aligned}
    & \mathbf{{M}}^{g}_{[N_s^g-N_m:N_s^g]},\quad \mathrm{if} \  N_s^g > N_m \\
    & \mathbf{{M}}^{g}, \quad \mathrm{otherwise}
    \end{aligned}
    \right. 
\end{equation}
where $N_m$ and $N_s^g$ denote the maximum storage limit and the current storage volume of each memory matrix, respectively. The algorithm of the memory storing and updating operation is summarized in the appendix.

\textbf{Strategy Prediction}. 
In order to use the strategy pattern information in the memory bank, the model requires selecting a proper strategy category based on the dialogue context. To achieve these, we leverage a strategy predictor, which aims to capture information relevant to strategy decisions in the context.

The strategy predictor is composed of a Transformer-based encoder and a classification module. The encoder first encodes the dialogue context into a strategy predict representation. It is worth noting that we adopt independent representations for strategy prediction and response generation tasks considering the fact that jointly optimal solutions may not always exist for different tasks. Subsequently, the classification module maps the vector as a $G$ dimension vector, which is regarded as the probability distribution of the $G$ strategy types. Formally, the strategy prediction can be written as follows,
\begin{equation} 
 \left\{
 \begin{aligned}
    & \mathbf{s} =\operatorname{MaxPooling} (\operatorname{Enc}_s(I)), \\
    &  \hat{g} =\operatorname{argmax}( \sigma(\operatorname{MLP}(\mathbf{s})),
  \end{aligned}
 \right.
 \end{equation}
where $\operatorname{Enc}_s$ is a Transformer-based encoder. The strategy prediction representation is denoted as $\mathbf{s} \in \mathbb{R}^d$. $\operatorname{MLP}$ and $\sigma(\cdot)$ are a multi-layer perceptron and the Sigmoid function, respectively. The argmax operation is used to obtain the predicted strategy category $\hat{g}$. 
We use the following objective to optimize the strategy prediction task, 
\begin{equation}
    \mathcal{L}_s= -\log {p}\left({g}|\mathbf{s} \right).
\end{equation}

\textbf{Memory-enhanced Encoding.} 
After predicting the strategy category $\hat{g}$, instead of directly using $\hat{g}$ as an indicator to constrain the response generation, we integrate the aforementioned corresponding memory bank matrix $\mathbf{M}^{g}$ and the context representation, so as to fully exploit the abundant pattern information of the particular strategy. 

Empirically, we fuse the matrix and the context representation with a multi-head cross-attention module~\cite{DBLP:conf/nips/VaswaniSPUJGKP17} as follows,
\begin{equation}
\mathbf{m}=\operatorname{MaxPooling}(\operatorname{CrossAtt}(\mathbf{H}, \mathbf{M}^g)),
\end{equation}
where $\mathbf{H}$ and $\mathbf{M}^g$ act as the \textit{query} and the \textit{key-value} pair in the cross-attention, respectively, $\mathbf{m} \in \mathbb{R}^{d}$ denotes the memory-enhanced strategy modeling feature.

\subsection{Response Decoding}
In order to generate the emotional supportive response, we input the encoded features, \ie the memory-enhanced strategy modeling feature $\mathbf{m}$ and the knowledge-enriched dialogue context embedding $\mathbf{H}$, into the Transformer decoder. 
Formally, we deploy the decoding process, which predicts the conditional probability distribution over each token in the target response in an auto-regressive manner as follows,
\begin{equation}
 	{p} \left({\hat{r}}_{l} \mid {\hat{r}}_{<l}, \mathbf{E} \right) = \operatorname{Dec}(\hat{r}_{<l},\mathbf{E}),
\end{equation}
where $ \mathbf{E}=[\mathbf{m};\mathbf{H}]$. $\hat{r}_{<l}$ refers to the previous generated $l-1$ tokens of the target response. $\operatorname{Dec(\cdot)}$ denotes the decoder module. 
Notably, to avoid the accumulated error, we replace $\hat{r}_{<l}$ by ${r}_{<l}$ in the training phase. For optimization, we introduce the standard cross-entropy loss function for response generation as follows,
\begin{equation}
         \mathcal{L}_g= -\frac{1}{N_r}\sum_{l=1}^{N_r} \log {p} \left(r_l \mid r_{<l},\mathbf{E} \right), \\
\end{equation}
where $N_r$ denotes the length of the target response.
\subsection{Model Training}
Ultimately, we combine all the loss functions for optimizing our model as follows,
\begin{equation}
     \mathcal{L}= \mathcal{L}_g + \lambda_1  \mathcal{L}_s + \lambda_2  \mathcal{L}_r,
\label{eq:loss}
\end{equation}
where $\mathcal{L}$ is the final training objective and $\lambda_1$ and $\lambda_2$ are the non-negative hyperparameters used for balancing the effect of each loss function on the entire training process.  
The overall algorithm of the optimization is briefly summarized in the appendix.

\section{Experiments}
\subsection{Dataset}
We conducted experiments on the ESConv dataset~\cite{DBLP:conf/acl/LiuZDSLYJH20}. Each sample in the dataset is a dialogue between a help-seeker and a supporter. In addition to the context, it also contains rich information, \eg the situation that the help-seeker's is facing.
It also provides the annotation of the strategy category used in every supporter's response. The dataset contains 1,300 long conversations and overall 38,350 utterances, with an average of 29.5 utterances in each dialogue. For the data split, we followed the same setting of previous work~\cite{DBLP:conf/acl/LiuZDSLYJH20,DBLP:conf/emnlp/ChengLLWZLL022}. 

\begin{table*}[htbp]
  \centering
  \caption{Performance comparison under automatic evaluations. The best results are highlighted in bold.}
    \resizebox{0.7\textwidth}{!}{
    \begin{tabular}{l|cccccccc}
    \toprule
       \textbf{Model}   & \textbf{PPL} & \textbf{B-1} & \textbf{B-2} & \textbf{B-3} & \textbf{B-4} & \textbf{R-L} & \textbf{MT} & \textbf{CIDEr} \\
    \midrule
    MoEL  & 264.11 & 19.04 & 6.47  & 2.91  & 1.51  & 15.95 & 7.96  & 10.95 \\
    MIME  & 69.28 & 15.24 & 5.56  & 2.64  & 1.50  & 16.12 & 6.43  & 10.66 \\
    DialoGPT-Joint & 15.71 & 17.39 & 5.59  & 2.03  & 1.18  & 16.93 & 7.55  & 11.86 \\
    BlenderBot-Joint & 16.79 & 17.62 & 6.91  & 2.81  & 1.66  & 17.94 & 7.54  & 18.04 \\
    MISC  & 16.16 & -     & 7.31  & -     & 2.20  & 17.91 & \textbf{11.05} & - \\
    GLHG  & 15.67 & 19.66 & 7.57  & 3.74  & 2.13  & 16.37 & -     & - \\
    FADO  & 15.72 & -     & 8.00  & 4.00  & 2.32  & 17.53 & -     & - \\
    PoKE  & 15.84 & 18.41 & 6.79  & 3.24  & 1.78  & 15.84 & -     & - \\
    MultiESC & 15.41 & 21.65 & 9.18  & 4.99  & 3.09  & 20.41 & 8.84  & 29.98 \\
    \midrule
    \textbf{MODERN} & \textbf{14.99} & \textbf{23.19} & \textbf{10.13} & \textbf{5.53} & \textbf{3.39} & \textbf{20.86} & 9.26  & \textbf{30.08} \\
    \bottomrule
    \end{tabular}%
    }
  \label{tab:main}%
\end{table*}%

\subsection{Implementation Details}
Following the setting of the previous study~\cite{DBLP:conf/emnlp/ChengLLWZLL022}, we also utilized encoder and decoder of the pretrained BART\footnote{\url{https://huggingface.co/facebook/bart-base}.} provided by HuggingFace~\cite{DBLP:journals/corr/abs-1910-03771} to initialize the parameters of the context encoder, the strategy predictor, and the decoder, respectively. 
To form a mini-batch, the input sequence length $L$ is unified to 512. The hidden dimension $d$ is 768. The category number $G$ equals $8$. The maximum memory storage $N_m$ is set as $64$. $K$ equals 20.  $\lambda_1$ and  $\lambda_2$ are $0.3$ and $0.1$, respectively. 
The batch size is 16. We use the PPL metric on the validation set to monitor the training progress. Empirically, it takes around 15 epochs to get the peak performance. 
More implementation details can be seen in the appendix. 


\subsection{Evaluation Metrics}
For the comprehensive evaluation, we conducted both automatic and human evaluations. 
\paragraph{Automatic Evaluation} For the automatic evaluation, we adopt several mainstream metrics commonly used in dialogue response tasks, \ie PPL (Perplexity), BLEU-\{1,2,3,4\} (B-\{1,2,3,4\})~\cite{DBLP:conf/acl/PapineniRWZ02}, ROUGE-L (R-L)~\cite{lin2004rouge}, METEOR (MT)~\cite{lavie-agarwal-2007-meteor}, and CIDEr~\cite{DBLP:conf/cvpr/VedantamZP15}.

\paragraph{Human Evaluation} 
In this part, we first randomly select $70$ testing samples for evaluation. Then, we employ $2$ volunteers to manually choose which response outperforms the other one.  Every sample is annotated twice. For each case, the volunteers need to compare the generated texts of different models according to the following four dimensions: 1)~\textbf{Fluency}: which response is more fluent, correct, and coherent in grammar and syntax; 2)~\textbf{Relevance}: which response talks more relevantly regarding current dialogue context; 3)~\textbf{Empathy}: which response is better to react with appropriate emotion according to the user's emotional state; 4)~\textbf{Information}: which response provides more suggestive information to help solve the problem. To further control the quality of the evaluation, we also invite an inspector to randomly sample and double-check 10\% rating results.

\begin{table}[tbp]
  \centering
  \caption{The human evaluation results in four dimensions.}
    \resizebox{0.75\linewidth}{!}{
    \begin{tabular}{c|c|ccc}
    \toprule
    \textbf{Comparisons}   & \textbf{Aspects} & \textbf{Win}   & \textbf{Tie}  & \textbf{Lose}  \\
    \midrule
    \multirow{4}[2]{*}{vs. FADO} & Fluency & 37.0 & 24.0  & 9.0 \\
    & Relevance & 35.5  & 20.5  & 14.0 \\
    & Empathy & 34.0  & 16.5  & 19.5 \\
    & Information & 37.5  & 17.0  & 15.5 \\
    \midrule
    \multirow{4}[2]{*}{vs. MultiESC} & Fluency & 30.5 & 24.5 & 15.0 \\ 
    & Relevance & 32.5  & 20.5  & 17.0 \\
    & Empathy & 35.0  & 19.5  & 15.5 \\ 
    & Information & 37.5  & 21.0  & 11.5 \\
\bottomrule
    \end{tabular} }
  \label{tab:humaneval}
\end{table}%

\subsection{Model Comparison \& Analysis}
To validate the effectiveness of our model, we compare it with several state-of-the-art baselines: \textbf{MoEL}~\cite{DBLP:conf/emnlp/LinMSXF19}, \textbf{MIME}~\cite{DBLP:conf/emnlp/MajumderHPLGGMP20}, \textbf{DialoGPT-Joint}~\cite{DBLP:conf/acl/LiuZDSLYJH20}, \textbf{BlenderBot-Joint}~\cite{DBLP:conf/acl/LiuZDSLYJH20},  \textbf{MISC}~\cite{DBLP:conf/acl/TuLC0W022}, \textbf{GLHG}~\cite{DBLP:conf/ijcai/00080XXSL22}, \textbf{PoKE}~\cite{DBLP:journals/corr/abs-2210-12640}, \textbf{FADO}~\cite{DBLP:journals/kbs/PengQHXL23}, \textbf{MultiESC}~\cite{DBLP:conf/emnlp/ChengLLWZLL022}. 
More details about them are described in the appendix. 


\begin{table}[t]
  \centering
  \caption{Experimental results of ablation study.}
    \resizebox{\linewidth}{!}{
    \begin{tabular}{l|cccccccc}
    \toprule
       \textbf{Model}   & \textbf{PPL} & \textbf{B-1} & \textbf{B-2} & \textbf{B-3} & \textbf{B-4} & \textbf{R-L} & \textbf{MT} & \textbf{CIDEr} \\
    \midrule
    w/o-$\mathcal{L}_s$ & 15.88 & 20.25 & 8.61  & 4.68  & 2.91  & 20.11 & 8.44  & 24.32 \\
    w/o-$\mathcal{L}_r$ & 15.32 & 21.69 & 9.31  & 5.06  & 3.16  & 20.57 & 8.87  & 28.91 \\
    w/o-Mem & 15.84 & 21.24 & 8.90  & 4.70  & 2.87  & 20.21 & 8.63  & 27.55 \\
    w/o-Emo & 15.91 & 20.35 & 8.46  & 4.48  & 2.72  & 19.79 & 8.27  & 24.01 \\
    w/o-KG & 15.58 & 21.56 & 9.08  & 4.82  & 2.96  & 19.98 & 8.74  & 26.83 \\
    \midrule
    \textbf{MODERN } & \textbf{14.99} & \textbf{23.19} & \textbf{10.13} & \textbf{5.53} & \textbf{3.39} & \textbf{20.86} & \textbf{9.26} & \textbf{30.08} \\
    \bottomrule
    \end{tabular}%
    }
  \label{tab:ablation}%
\end{table}%

\paragraph{Automatic Evaluation} 
We compared our model with the above baselines using automatic metrics and results are reported in Table~\ref{tab:main}. As we can see, 1) MODERN outperforms the baselines in most metrics, which is a powerful proof of the effectiveness of the proposed method. 
2) The models with BART backbone (\ie MultiESC and MODERN) surpass those baselines with BlenderBot~\cite{roller-etal-2021-recipes} backbone (\ie BlenderBot-Joint, MISC, GLHG, FADO, and PokE) across most of the metrics despite the latter being pretrained on empathetic-related data. One possible explanation is that the ESConv task requires sophisticated linguistic knowledge (e.g. correct grammar and wording appropriate to the current situation) in addition to empathetic ability.
3) Our MODERN consistently exceeds MultiESC with the same BART backbone. This suggests that BART model with large-scale pretraining still requires strategy pattern information and knowledge (\ie emotion and concepts) to further facilitate supportive response generation. 

\paragraph{Human Evaluation} For human evaluation, we report the comparison results between our model and the two best baselines (\ie FADO and MultiESC) in Table~\ref{tab:humaneval}. 
In particular, for each pair of model comparisons and each metric, we show the number of samples where our model achieves better (denoted as ``Win''), equal (denoted as ``Tie''), and worse performance (denoted as ``Lose'') compared with the baselines.
As seen, MODERN outperforms all baselines across different evaluation metrics, as the number of ``Win'' cases is always significantly larger than that of ``Lose'' cases in each pair of model comparisons, which is consistent with the results in Table~\ref{tab:main}. 
In addition, the number of ``Win'' cases is the largest for the Information metric compared with other metrics, which demonstrates that integrating context-related concepts can supply meaningful information for emotional support. 


\begin{figure*}[t]
    \centering
    \includegraphics[width=0.95\textwidth]{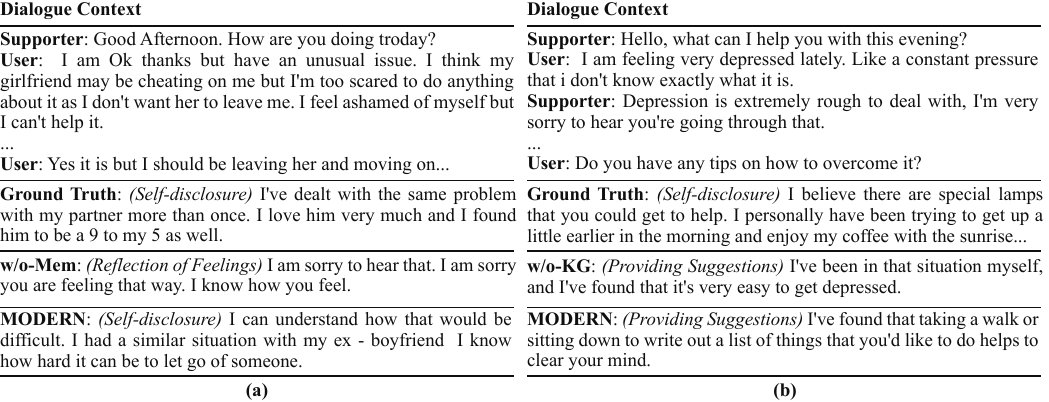}
    \caption{Intuitive comparison of the MODERN and two derivatives.} 
    \label{fig:case}
\end{figure*}
\subsection{Ablation Study}
We compare the original MODERN model with the following derivatives to demonstrate that all the designed modules are essential for the ESConv task. 
1) \textbf{w/o-$\mathcal{L}_s$}. To show the benefit of constraint for strategy prediction, we removed the corresponding loss function by setting $\lambda_1 = 0$ in Eqn.(\ref{eq:loss}).    
2) \textbf{w/o-$\mathcal{L}_r$}. To show the effect of the auxiliary
strategy classification task, we removed the corresponding loss function by setting $\lambda_2 = 0$ in Eqn.(\ref{eq:loss}).    
3) \textbf{w/o-Mem}. In this derivative, we disabled the memory module, which stores pattern representations of different strategies. 
4) \textbf{w/o-Emo}. In this derivative, we removed the change-aware emotion detection module. 
And 5) \textbf{w/o-KG}. We discarded the context-relate concepts reasoning and selection component. 

We provided the ablation study results on the ESConv dataset in Table~\ref{tab:ablation} in terms of all metrics. From this table, we have the following observations: 1) Our MODERN consistently outperforms w/o-Mem, especially on the BLUE metrics (B-\{1,2,3,4\}), which suggests that the memory-enhanced strategy modeling module can provide sufficient linguistic pattern references and hence boost the performance of generating responses in accordance with specific strategy categories. 2) MODERN exceeds w/o-$\mathcal{L}_s$ across all metrics. This verifies it is indispensable to constrain the model to predict and respond under proper strategies. 3) w/o-$\mathcal{L}_r$ obtains a slightly worse result than MODERN, which demonstrates the strategy classification auxiliary task indeed helps with guiding the pattern representation learning. And 4) w/o-Emo and w/o-KG both perform worse than MODERN, which demonstrates the importance of change-aware emotion and context-related concepts. 
Notably, w/o-KG surpasses w/o-Emo. One possible explanation is that being aware of the dynamic emotional changes during the conversation facilitates the model to provide empathy and emotional support accordingly.

\subsection{Case Study}
We illustrate several conversations in the test set to get an intuitive understanding  of our model in Figure~\ref{fig:case}. 
We showed two samples and compare the responses generated by MODERN and two derivatives, \ie w/o-Mem and w/o-KG. 
As can be seen in case (a), MODERN fulfills to respond with the strategy of \textit{Self-disclosure} and generates a contextually appropriate response. Being equipped with a memory-enhanced strategy modeling module, MODERN shares a similar experience closely related to the seeker's problem \ie \textit{relationship issue}. Whereas the w/o-Mem model generates a plain and monotonous response, which is not very relevant to the user's current issue. 
The other case (b) demonstrates how MODERN reaps benefits from external knowledge. Based on the situation that the seeker mentions \textit{feeling depressed}, MODERN leverages the context-related concepts and associates this emotion status with practical suggestions \ie \textit{taking a walk or sitting down to write...} effectively. While without relevant knowledge, the response generated by the w/o-KG derivative is relatively vague and less specific, which is deficient to benefit the user's situation. 

\section{Conclusions}
In this paper, we propose a novel knowledge-enhanced Memory mODEl for emotional suppoRt coNversation, dubbed MODERN, which can perceive fine-grained emotional changes in the conversation, utilize the concepts from knowledge graph to facilitate generating responses with practical suggestions, and model concrete strategy semantic patterns with memory bank mechanism.  
The experimental results show that our model surpasses all the state-of-the-art methods. 
In addition, the ablation study demonstrates the effectiveness of each component of our model. 

\section*{Limitations}
The ESConv task requires the dialogue agent to reveal some information about itself. For example, one of the strategies called \textit{Self-disclosure}, expects the agent to cite their own experience. However, in our experiments, we observed that the current model often struggles to maintain a consistent personality. We speculate that this may be due to the fact that the supporter role in the full training sample is provided by multiple individuals, and thus there is no uniform character experience and story, which leads to the problem of inconsistent personal experiences during the conversation. We believe that how to make the dialogue agent show coherent and unified personal information and experiences in the ESConv task deserves the attention of future work.





\bibliography{anthology,reference}

\begin{thebibliography}{43}
\expandafter\ifx\csname natexlab\endcsname\relax\def\natexlab#1{#1}\fi

\bibitem[{Chang and Chen(2021)}]{chang2021relationships}
Ching-Wen Chang and Fang-pei Chen. 2021.
\newblock Relationships of family emotional support and negative family
  interactions with the quality of life among chinese people with mental
  illness and the mediating effect of internalized stigma.
\newblock \emph{Psychiatric Quarterly}, 92(1):375--387.

\bibitem[{Chen et~al.(2022)Chen, Li, and Yang}]{DBLP:conf/naacl/ChenLY22}
Mao~Yan Chen, Siheng Li, and Yujiu Yang. 2022.
\newblock Emphi: Generating empathetic responses with human-like intents.
\newblock In \emph{Proceedings of the Conference of the North American Chapter
  of the Association for Computational Linguistics: Human Language
  Technologies}, pages 1063--1074. {ACL}.

\bibitem[{Cheng et~al.(2022)Cheng, Liu, Li, Wang, Zhao, Liu, Liang, and
  Zheng}]{DBLP:conf/emnlp/ChengLLWZLL022}
Yi~Cheng, Wenge Liu, Wenjie Li, Jiashuo Wang, Ruihui Zhao, Bang Liu, Xiaodan
  Liang, and Yefeng Zheng. 2022.
\newblock Improving multi-turn emotional support dialogue generation with
  lookahead strategy planning.
\newblock In \emph{Proceedings of the Conference on Empirical Methods in
  Natural Language Processing}, pages 3014--3026. ACL.

\bibitem[{Ding et~al.(2022)Ding, Liu, Zhang, and Yang}]{Ding2022}
Yue Ding, Jingjing Liu, Xiaochen Zhang, and Zhi Yang. 2022.
\newblock Dynamic tracking of state anxiety via multi-modal data and machine
  learning.
\newblock \emph{Frontiers in psychiatry}, 13.

\bibitem[{Green et~al.(2022)Green, Faizi, Jalal, and
  Zadran}]{green2022emotional}
Zane~Asher Green, Farkhonda Faizi, Rahmatullah Jalal, and Zarifa Zadran. 2022.
\newblock Emotional support received moderates academic stress and mental
  well-being in a sample of afghan university students amid covid-19.
\newblock \emph{International Journal of Social Psychiatry}, 68(8):1748--1755.

\bibitem[{Greene and Burleson(2003)}]{Burleson2003}
J.O. Greene and B.R. Burleson. 2003.
\newblock \emph{Handbook of Communication and Social Interaction Skills}.
\newblock American Psychological Association.

\bibitem[{High and Steuber(2014)}]{HighSteuber2014}
Andrew~C. High and Keli~Ryan Steuber. 2014.
\newblock An examination of support (in)adequacy: Types, sources, and
  consequences of social support among infertile women.
\newblock \emph{Communication Monographs}, 81.

\bibitem[{Hill(2009)}]{hill2009helping}
Clara~E Hill. 2009.
\newblock \emph{Helping skills: Facilitating, exploration, insight, and
  action}.
\newblock American Psychological Association.

\bibitem[{Jing et~al.(2023)Jing, Song, Lin, Zhao, Zhou, and Nie}]{stylized}
Liqiang Jing, Xuemeng Song, Xuming Lin, Zhongzhou Zhao, Wei Zhou, and Liqiang
  Nie. 2023.
\newblock Stylized data-to-text generation: A case study in the e-commerce
  domain.
\newblock \emph{ACM Trans. Inf. Syst.}

\bibitem[{Kim et~al.(2022)Kim, Ahn, Kim, and Lee}]{DBLP:conf/naacl/KimAKL22}
Wongyu Kim, Youbin Ahn, Donghyun Kim, and Kyong{-}Ho Lee. 2022.
\newblock Emp-rft: Empathetic response generation via recognizing feature
  transitions between utterances.
\newblock In \emph{Proceedings of the Conference of the North American Chapter
  of the Association for Computational Linguistics: Human Language
  Technologies}, pages 4118--4128. {ACL}.

\bibitem[{Lavie and Agarwal(2007)}]{lavie-agarwal-2007-meteor}
Alon Lavie and Abhaya Agarwal. 2007.
\newblock \href {https://aclanthology.org/W07-0734} {{METEOR}: An automatic
  metric for {MT} evaluation with high levels of correlation with human
  judgments}.
\newblock In \emph{Proceedings of the Second Workshop on Statistical Machine
  Translation}, pages 228--231, Prague, Czech Republic. Association for
  Computational Linguistics.

\bibitem[{Lewis et~al.(2020)Lewis, Liu, Goyal, Ghazvininejad, Mohamed, Levy,
  Stoyanov, and Zettlemoyer}]{DBLP:conf/acl/LewisLGGMLSZ20}
Mike Lewis, Yinhan Liu, Naman Goyal, Marjan Ghazvininejad, Abdelrahman Mohamed,
  Omer Levy, Veselin Stoyanov, and Luke Zettlemoyer. 2020.
\newblock {BART:} denoising sequence-to-sequence pre-training for natural
  language generation, translation, and comprehension.
\newblock In \emph{Proceedings of the Annual Meeting of the Association for
  Computational Linguistics}, pages 7871--7880. ACL.

\bibitem[{Li et~al.(2020)Li, Chen, Ren, Ren, Tu, and
  Chen}]{DBLP:conf/coling/LiCRRTC20}
Qintong Li, Hongshen Chen, Zhaochun Ren, Pengjie Ren, Zhaopeng Tu, and Zhumin
  Chen. 2020.
\newblock Empdg: Multi-resolution interactive empathetic dialogue generation.
\newblock In \emph{Proceedings of the International Conference on Computational
  Linguistics}, pages 4454--4466. International Committee on Computational
  Linguistics.

\bibitem[{Li et~al.(2022)Li, Li, Ren, Ren, and Chen}]{DBLP:conf/aaai/LiLRRC22}
Qintong Li, Piji Li, Zhaochun Ren, Pengjie Ren, and Zhumin Chen. 2022.
\newblock Knowledge bridging for empathetic dialogue generation.
\newblock In \emph{The Conference on Artificial Intelligence, Conference on
  Innovative Applications of Artificial Intelligence, The Symposium on
  Educational Advances in Artificial Intelligence}, pages 10993--11001. {AAAI}
  Press.

\bibitem[{Lin(2004)}]{lin2004rouge}
Chin-Yew Lin. 2004.
\newblock Rouge: A package for automatic evaluation of summaries.
\newblock In \emph{Text summarization branches out}, pages 74--81.

\bibitem[{Lin et~al.(2019)Lin, Madotto, Shin, Xu, and
  Fung}]{DBLP:conf/emnlp/LinMSXF19}
Zhaojiang Lin, Andrea Madotto, Jamin Shin, Peng Xu, and Pascale Fung. 2019.
\newblock Moel: Mixture of empathetic listeners.
\newblock In \emph{Proceedings of the Conference on Empirical Methods in
  Natural Language Processing and the International Joint Conference on Natural
  Language Processing}, pages 121--132. {ACL}.

\bibitem[{Liu and Sundar(2018)}]{liu2018should}
Bingjie Liu and S~Shyam Sundar. 2018.
\newblock Should machines express sympathy and empathy? experiments with a
  health advice chatbot.
\newblock \emph{Cyberpsychology, Behavior, and Social Networking},
  21(10):625--636.

\bibitem[{Liu et~al.(2021{\natexlab{a}})Liu, Zheng, Demasi, Sabour, Li, Yu,
  Jiang, and Huang}]{DBLP:conf/acl/LiuZDSLYJH20}
Siyang Liu, Chujie Zheng, Orianna Demasi, Sahand Sabour, Yu~Li, Zhou Yu, Yong
  Jiang, and Minlie Huang. 2021{\natexlab{a}}.
\newblock Towards emotional support dialog systems.
\newblock In \emph{Proceedings of the Annual Meeting of the Association for
  Computational Linguistics and the International Joint Conference on Natural
  Language Processing}, pages 3469--3483. {ACL}.

\bibitem[{Liu et~al.(2018)Liu, Bi, Gao, Liu, Yao, and
  Shi}]{liu-etal-2018-towards-less}
Yahui Liu, Wei Bi, Jun Gao, Xiaojiang Liu, Jian Yao, and Shuming Shi. 2018.
\newblock \href {https://doi.org/10.18653/v1/D18-1297} {Towards less generic
  responses in neural conversation models: A statistical re-weighting method}.
\newblock In \emph{Proceedings of the 2018 Conference on Empirical Methods in
  Natural Language Processing}, pages 2769--2774, Brussels, Belgium.
  Association for Computational Linguistics.

\bibitem[{Liu et~al.(2021{\natexlab{b}})Liu, Maier, Minker, and
  Ultes}]{DBLP:conf/iwsds/Liu0MU21}
Ye~Liu, Wolfgang Maier, Wolfgang Minker, and Stefan Ultes. 2021{\natexlab{b}}.
\newblock Empathetic dialogue generation with pre-trained roberta-gpt2 and
  external knowledge.
\newblock In \emph{Conversational {AI} for Natural Human-Centric Interaction
  International Workshop on Spoken Dialogue System Technology}, volume 943 of
  \emph{Lecture Notes in Electrical Engineering}, pages 67--81. Springer.

\bibitem[{Loshchilov and Hutter(2017)}]{DBLP:journals/corr/abs-1711-05101}
Ilya Loshchilov and Frank Hutter. 2017.
\newblock Fixing weight decay regularization in adam.
\newblock \emph{CoRR}, abs/1711.05101.

\bibitem[{Majumder et~al.(2020)Majumder, Hong, Peng, Lu, Ghosal, Gelbukh,
  Mihalcea, and Poria}]{DBLP:conf/emnlp/MajumderHPLGGMP20}
Navonil Majumder, Pengfei Hong, Shanshan Peng, Jiankun Lu, Deepanway Ghosal,
  Alexander~F. Gelbukh, Rada Mihalcea, and Soujanya Poria. 2020.
\newblock {MIME:} mimicking emotions for empathetic response generation.
\newblock In \emph{Proceedings of the Conference on Empirical Methods in
  Natural Language Processing}, pages 8968--8979. {ACL}.

\bibitem[{Mohammad(2018)}]{nrcvadcite}
Saif~M. Mohammad. 2018.
\newblock Obtaining reliable human ratings of valence, arousal, and dominance
  for 20, 000 english words.
\newblock In \emph{Proceedings of the Annual Meeting of the Association for
  Computational Linguistics}, pages 174--184. ACL.

\bibitem[{Papineni et~al.(2002)Papineni, Roukos, Ward, and
  Zhu}]{DBLP:conf/acl/PapineniRWZ02}
Kishore Papineni, Salim Roukos, Todd Ward, and Wei{-}Jing Zhu. 2002.
\newblock Bleu: a method for automatic evaluation of machine translation.
\newblock In \emph{Proceedings of the Annual Meeting of the Association for
  Computational Linguistics}, pages 311--318. {ACL}.

\bibitem[{Parkin et~al.(2014)Parkin, de~Looy, and Farrand}]{parkin2014greater}
Tracey Parkin, Anne de~Looy, and Paul Farrand. 2014.
\newblock Greater professional empathy leads to higher agreement about
  decisions made in the consultation.
\newblock \emph{Patient Education and Counseling}, 96(2):144--150.

\bibitem[{Paszke et~al.(2019)Paszke, Gross, Massa, Lerer, Bradbury, Chanan,
  Killeen, Lin, Gimelshein, Antiga, Desmaison, K{\"{o}}pf, Yang, DeVito,
  Raison, Tejani, Chilamkurthy, Steiner, Fang, Bai, and
  Chintala}]{DBLP:conf/nips/PaszkeGMLBCKLGA19}
Adam Paszke, Sam Gross, Francisco Massa, Adam Lerer, James Bradbury, Gregory
  Chanan, Trevor Killeen, Zeming Lin, Natalia Gimelshein, Luca Antiga, Alban
  Desmaison, Andreas K{\"{o}}pf, Edward~Z. Yang, Zachary DeVito, Martin Raison,
  Alykhan Tejani, Sasank Chilamkurthy, Benoit Steiner, Lu~Fang, Junjie Bai, and
  Soumith Chintala. 2019.
\newblock Pytorch: An imperative style, high-performance deep learning library.
\newblock In \emph{Advances in Neural Information Processing Systems: Annual
  Conference on Neural Information Processing Systems}, pages 8024--8035.

\bibitem[{Peng et~al.(2022)Peng, Hu, Xing, Xie, Sun, and
  Li}]{DBLP:conf/ijcai/00080XXSL22}
Wei Peng, Yue Hu, Luxi Xing, Yuqiang Xie, Yajing Sun, and Yunpeng Li. 2022.
\newblock Control globally, understand locally: {A} global-to-local
  hierarchical graph network for emotional support conversation.
\newblock In \emph{Proceedings of the International Joint Conference on
  Artificial Intelligence}, pages 4324--4330. ijcai.org.

\bibitem[{Peng et~al.(2023)Peng, Qin, Hu, Xie, and
  Li}]{DBLP:journals/kbs/PengQHXL23}
Wei Peng, Ziyuan Qin, Yue Hu, Yuqiang Xie, and Yunpeng Li. 2023.
\newblock {FADO:} feedback-aware double controlling network for emotional
  support conversation.
\newblock \emph{Knowledge-Based Systems}, 264:110340.

\bibitem[{Rashkin et~al.(2019)Rashkin, Smith, Li, and
  Boureau}]{DBLP:conf/acl/RashkinSLB19}
Hannah Rashkin, Eric~Michael Smith, Margaret Li, and Y{-}Lan Boureau. 2019.
\newblock Towards empathetic open-domain conversation models: {A} new benchmark
  and dataset.
\newblock In \emph{Proceedings of the Conference of the Association for
  Computational Linguistics}, pages 5370--5381. {ACL}.

\bibitem[{Reynolds and Scott(1999)}]{reynolds1999empathy}
William~J Reynolds and Brain Scott. 1999.
\newblock Empathy: a crucial component of the helping relationship.
\newblock \emph{Journal of psychiatric and mental health nursing},
  6(5):363--370.

\bibitem[{Roller et~al.(2021)Roller, Dinan, Goyal, Ju, Williamson, Liu, Xu,
  Ott, Smith, Boureau, and Weston}]{roller-etal-2021-recipes}
Stephen Roller, Emily Dinan, Naman Goyal, Da~Ju, Mary Williamson, Yinhan Liu,
  Jing Xu, Myle Ott, Eric~Michael Smith, Y-Lan Boureau, and Jason Weston. 2021.
\newblock \href {https://doi.org/10.18653/v1/2021.eacl-main.24} {Recipes for
  building an open-domain chatbot}.
\newblock In \emph{Proceedings of the 16th Conference of the European Chapter
  of the Association for Computational Linguistics: Main Volume}, pages
  300--325, Online. Association for Computational Linguistics.

\bibitem[{Shen and Feng(2020)}]{DBLP:conf/acl/ShenF20}
Lei Shen and Yang Feng. 2020.
\newblock {CDL:} curriculum dual learning for emotion-controllable response
  generation.
\newblock In \emph{Proceedings of the Annual Meeting of the Association for
  Computational Linguistics}, pages 556--566. {ACL}.

\bibitem[{Sohn et~al.(2015)Sohn, Lee, and Yan}]{DBLP:conf/nips/SohnLY15}
Kihyuk Sohn, Honglak Lee, and Xinchen Yan. 2015.
\newblock Learning structured output representation using deep conditional
  generative models.
\newblock In \emph{Advances in Neural Information Processing Systems 28: Annual
  Conference on Neural Information Processing Systems}, pages 3483--3491.

\bibitem[{Speer et~al.(2017)Speer, Chin, and Havasi}]{DBLP:conf/aaai/SpeerCH17}
Robyn Speer, Joshua Chin, and Catherine Havasi. 2017.
\newblock Conceptnet 5.5: An open multilingual graph of general knowledge.
\newblock In \emph{Proceedings of the Thirty-First {AAAI} Conference on
  Artificial Intelligence}, pages 4444--4451. {AAAI} Press.

\bibitem[{Tu et~al.(2022)Tu, Li, Cui, Wang, Wen, and
  Yan}]{DBLP:conf/acl/TuLC0W022}
Quan Tu, Yanran Li, Jianwei Cui, Bin Wang, Ji{-}Rong Wen, and Rui Yan. 2022.
\newblock {MISC:} {A} mixed strategy-aware model integrating {COMET} for
  emotional support conversation.
\newblock In \emph{Proceedings of the Annual Meeting of the Association for
  Computational Linguistics}, pages 308--319. {ACL}.

\bibitem[{Vaswani et~al.(2017)Vaswani, Shazeer, Parmar, Uszkoreit, Jones,
  Gomez, Kaiser, and Polosukhin}]{DBLP:conf/nips/VaswaniSPUJGKP17}
Ashish Vaswani, Noam Shazeer, Niki Parmar, Jakob Uszkoreit, Llion Jones,
  Aidan~N. Gomez, Lukasz Kaiser, and Illia Polosukhin. 2017.
\newblock Attention is all you need.
\newblock In \emph{Advances in Neural Information Processing Systems 30: Annual
  Conference on Neural Information Processing Systems}, pages 5998--6008.

\bibitem[{Vedantam et~al.(2015)Vedantam, Zitnick, and
  Parikh}]{DBLP:conf/cvpr/VedantamZP15}
Ramakrishna Vedantam, C.~Lawrence Zitnick, and Devi Parikh. 2015.
\newblock Cider: Consensus-based image description evaluation.
\newblock In \emph{Conference on Computer Vision and Pattern Recognition},
  pages 4566--4575. {IEEE}.

\bibitem[{Wei et~al.(2019)Wei, Lu, Mou, Zhou, Poupart, Li, and
  Jin}]{DBLP:conf/icassp/WeiLMZPLJ19}
Bolin Wei, Shuai Lu, Lili Mou, Hao Zhou, Pascal Poupart, Ge~Li, and Zhi Jin.
  2019.
\newblock Why do neural dialog systems generate short and meaningless replies?
  a comparison between dialog and translation.
\newblock In \emph{International Conference on Acoustics, Speech and Signal
  Processing}, pages 7290--7294. {IEEE}.

\bibitem[{Wolf et~al.(2019)Wolf, Debut, Sanh, Chaumond, Delangue, Moi, Cistac,
  Rault, Louf, Funtowicz, and Brew}]{DBLP:journals/corr/abs-1910-03771}
Thomas Wolf, Lysandre Debut, Victor Sanh, Julien Chaumond, Clement Delangue,
  Anthony Moi, Pierric Cistac, Tim Rault, R{\'{e}}mi Louf, Morgan Funtowicz,
  and Jamie Brew. 2019.
\newblock Huggingface's transformers: State-of-the-art natural language
  processing.
\newblock \emph{CoRR}, abs/1910.03771.

\bibitem[{Xu et~al.(2022)Xu, Meng, and
  Wang}]{DBLP:journals/corr/abs-2210-12640}
Xiaohan Xu, Xuying Meng, and Yequan Wang. 2022.
\newblock Poke: Prior knowledge enhanced emotional support conversation with
  latent variable.
\newblock \emph{CoRR}, abs/2210.12640.

\bibitem[{Zhang et~al.(2020)Zhang, Sun, Galley, Chen, Brockett, Gao, Gao, Liu,
  and Dolan}]{zhang-etal-2020-dialogpt}
Yizhe Zhang, Siqi Sun, Michel Galley, Yen-Chun Chen, Chris Brockett, Xiang Gao,
  Jianfeng Gao, Jingjing Liu, and Bill Dolan. 2020.
\newblock \href {https://doi.org/10.18653/v1/2020.acl-demos.30} {{DIALOGPT} :
  Large-scale generative pre-training for conversational response generation}.
\newblock In \emph{Proceedings of the 58th Annual Meeting of the Association
  for Computational Linguistics: System Demonstrations}, pages 270--278,
  Online. Association for Computational Linguistics.

\bibitem[{Zheng et~al.(2021)Zheng, Liu, Chen, Leng, and
  Huang}]{zheng-etal-2021-comae}
Chujie Zheng, Yong Liu, Wei Chen, Yongcai Leng, and Minlie Huang. 2021.
\newblock \href {https://doi.org/10.18653/v1/2021.findings-acl.72} {{C}o{MAE}:
  A multi-factor hierarchical framework for empathetic response generation}.
\newblock In \emph{Findings of the Association for Computational Linguistics:
  ACL-IJCNLP 2021}, pages 813--824, Online. Association for Computational
  Linguistics.

\bibitem[{Zhou et~al.(2018)Zhou, Huang, Zhang, Zhu, and
  Liu}]{DBLP:conf/aaai/ZhouHZZL18}
Hao Zhou, Minlie Huang, Tianyang Zhang, Xiaoyan Zhu, and Bing Liu. 2018.
\newblock Emotional chatting machine: Emotional conversation generation with
  internal and external memory.
\newblock In \emph{Proceedings of the {AAAI} Conference on Artificial
  Intelligence, the innovative Applications of Artificial Intelligence, and the
  {AAAI} Symposium on Educational Advances in Artificial Intelligence}, pages
  730--739. {AAAI} Press.

\end{thebibliography}
\bibliographystyle{acl_natbib}

\appendix
\section{Training Procedure.}
The overall algorithm of the optimization is briefly summarized in the algorithm~\ref{alg1}.

\begin{algorithm}[ht]
\caption{Training Procedure.} 
\label{alg1}
\begin{flushleft}
\hspace*{0.02in} {\bf Input:}
training set $\mathcal{P}$ for optimizing the model $\mathcal{F}$, hyperparameters \{$\lambda_1, \lambda_2$\}. \\  
\hspace*{0.02in} {\bf Output:} Parameters \textbf{ $\boldsymbol{\Theta}$}.\\
\end{flushleft}
\begin{algorithmic}[1]
\STATE Initialize parameters: {$\boldsymbol{\Theta}$} 
\STATE Initialize memory bank: {$\mathcal{{M}}$} 
\REPEAT
      \STATE Randomly sample a batch from $\mathcal{P}$.
      \FOR{each sample $(\mathcal{D},R,g,s)$}
    \STATE Add strategy pattern representation into the memory bank  $\mathcal{{M}}$ by Eqn.~(\ref{add}).
    \STATE Update the memory bank $\mathcal{{M}}$ by Eqn.~(\ref{update}).
    \ENDFOR 
    \STATE 	Update $\boldsymbol{\Theta}$ by optimizing the loss function in Eqn.~(\ref{eq:loss}).
\UNTIL{$\mathcal{F}$ converges.}

\end{algorithmic}
\end{algorithm}

\section{Implementation Details}

  During the generation stage, we use a maximum of 64 steps to decode the responses.
We adopt AdamW~\cite{DBLP:journals/corr/abs-1711-05101} optimizer with $\beta_1=0.9$ and $\beta_2=0.999$. The initial learning rate is $2e$-$5$ and a linear learning rate scheduler with 100 warm-up steps is used to reduce the learning rate progressively. 
For the framework, we use the PyTorch~\cite{DBLP:conf/nips/PaszkeGMLBCKLGA19} version 1.8.1 to implement the experiment codes. All experiments were conducted on an NVIDIA Tesla V100 32GB.

\section{Baselines}
\label{sec:appendix}

\textbf{MoEL}~\cite{DBLP:conf/emnlp/LinMSXF19}. This method adopts multiple decoders as different listeners for different emotions. The outputs of decoders are softly combined to generate the response. \\
     \textbf{MIME}~\cite{DBLP:conf/emnlp/MajumderHPLGGMP20}. This model shares the same architecture as MoEL and extends it to mimic the speaker's emotion.\\ 
     \textbf{DialoGPT-Joint}~\cite{DBLP:conf/acl/LiuZDSLYJH20}. This model is built on a pre-trained dialog agent DialoGPT~\cite{zhang-etal-2020-dialogpt}. It first predicts a strategy and prepends a special token before the response sentence to control the generation under that strategy. \\
     \textbf{BlenderBot-Joint}~\cite{DBLP:conf/acl/LiuZDSLYJH20}. This model adopts the same strategy prediction and generation scheme as DialoGPT. Differ from the former one, it is built on a pre-trained conversational response generation model named BlenderBot~\cite{roller-etal-2021-recipes}.\\
     \textbf{MISC}~\cite{DBLP:conf/acl/TuLC0W022}. This model also adopts BlenderBot as the backbone. It leverages common sense knowledge to enhance the understanding of the speaker's emotional state, and the response is generated conditioned on a mixture of strategy distribution. \\
     \textbf{GLHG}~\cite{DBLP:conf/ijcai/00080XXSL22}. This model has a global-to-local hierarchical graph structure. It models the global cause and the local intention of the speaker to provide more supportive responses. \\
      \textbf{PoKE}~\cite{DBLP:journals/corr/abs-2210-12640}.  This work utilized Conditional Variational Autoencoder~\cite{DBLP:conf/nips/SohnLY15} to model the mixed strategy. \\
       \textbf{FADO}~\cite{DBLP:journals/kbs/PengQHXL23}. This work devises a dual-level feedback strategy selector to encourage or penalize strategies during the strategy selection process. \\
        \textbf{MultiESC}~\cite{DBLP:conf/emnlp/ChengLLWZLL022} This work proposes lookahead heuristics to estimate the future strategy and capture users’ subtle emotional expressions with the NRC VAD lexicon~\cite{nrcvadcite} for  user state modeling.

\end{document}